\documentclass{article}

\usepackage{microtype}
\usepackage{graphicx}
\usepackage{subfigure}
\usepackage{booktabs} 

  \usepackage{amsmath}
  \usepackage{amsfonts}
  \usepackage{multirow}
  \usepackage{algorithm}
  \usepackage{algpseudocode}
  \usepackage{xcolor}
  \usepackage{amssymb}
  \usepackage{tikz}
\usepackage{hyperref}

\makeatletter\@namedef{ver@algorithmic.sty}{}\makeatother

\usepackage[accepted]{mlsys2025}
\makeatletter
\renewcommand{\Notice@String}{}
\makeatother

\mlsystitlerunning{Sonar-MoE: In-Backward Expert Updates for Memory-Efficient MoE Training}

  \newcommand{\sonar}{\textsc{Sonar-MoE}}
  
\begin{document}

\twocolumn[
\mlsystitle{\textsc{Sonar-MoE}: In-Backward Expert Updates for\texorpdfstring{\\}{ }Memory-Efficient Mixture-of-Experts Training}



\begin{mlsysauthorlist}
\mlsysauthor{Avni Mittal}{msft}
\mlsysauthor{Avinash Anand}{sit}
\mlsysauthor{Ashutosh Kumar}{midas}
\mlsysauthor{Dikshant Kukreja}{turboml}
\mlsysauthor{Kritarth Prasad}{sit}
\mlsysauthor{Chaitanya Lakhchaura}{iiitd}
\mlsysauthor{Rishitej Reddy Vyalla}{iiitd}
\mlsysauthor{Erik Cambria}{ntu}
\mlsysauthor{Zhengkui Wang}{sit}
\mlsysauthor{Rajiv Ratn Shah}{midas}
\end{mlsysauthorlist}

\mlsysaffiliation{msft}{Microsoft, India}
\mlsysaffiliation{sit}{Singapore Institute of Technology, Singapore}
\mlsysaffiliation{midas}{MIDAS Lab, Indian Institute of Technology Kanpur, Kanpur, India}
\mlsysaffiliation{turboml}{TurboML}
\mlsysaffiliation{iiitd}{Indraprastha Institute of Information Technology Delhi, New Delhi, India}
\mlsysaffiliation{ntu}{Nanyang Technological University, Singapore}

\mlsyscorrespondingauthor{Avni Mittal}{avni.mittal2002@gmail.com}

\mlsyskeywords{Machine Learning, MLSys}

\vskip 0.3in

\begin{abstract}
  Most parameters of a Mixture-of-Experts (MoE) model belong to its experts, and standard training keeps every expert gradient in GPU memory until the optimizer step. This memory grows with the number and size of the experts, so it rises as MoE models scale. We introduce \sonar{}, which updates each expert tensor during the backward pass and frees its gradient right away. It brings optimizer-in-backward from dense models to batched MoE expert tensors, so the GPU never holds more than one layer of expert gradients. When we fine-tune OLMoE-1B-7B and Granite-3.1-3B-A800M in bf16 with two expert-kernel backends, \sonar{} lowers peak memory by 19\% to 22\%. Its updates are bitwise identical to standard training without expert gradient clipping, and held-out loss stays within 0.003 of all baselines. Each step also takes 17\% to 23\% less time than standard training with global-norm clipping. Updating whole batched tensors rather than individual experts is what keeps \sonar{} fast, as it avoids a 20\% to 72\% slowdown while using the same memory. \sonar{} matches optimizer-in-backward over all parameters in memory and speed while still clipping the dense parameters.
\end{abstract}
]



\printAffiliationsAndNotice{\mlsysEqualContribution} 

  \section{Introduction}
\begin{figure*}[t]
  \centering
  \includegraphics[width=0.84\linewidth]{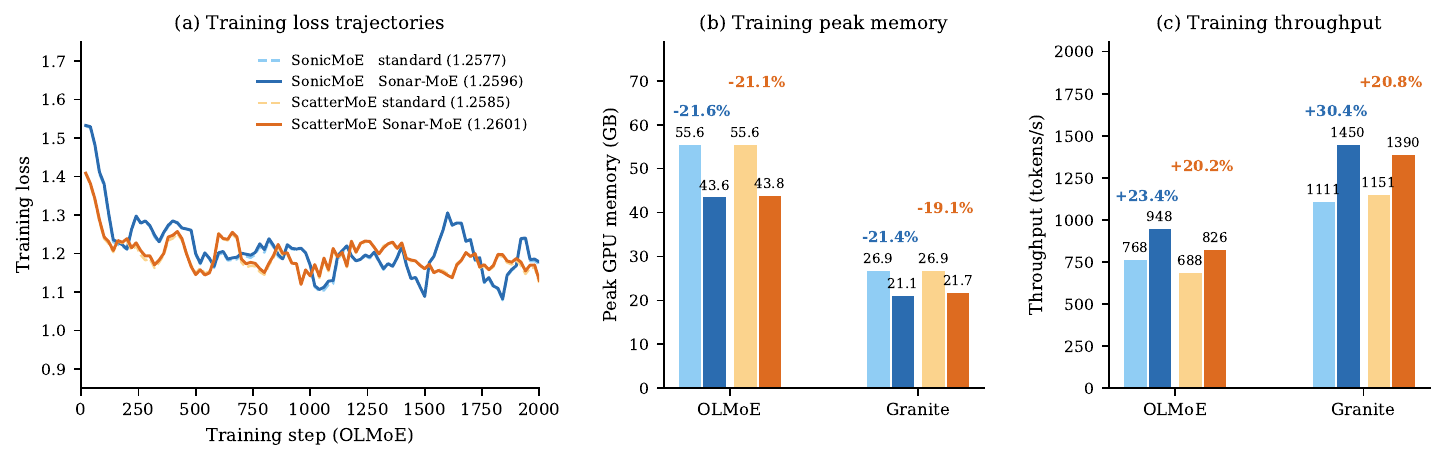}
  \caption{Alpaca fine-tuning of OLMoE-1B-7B and Granite-3.1-3B-A800M for 2{,}000 steps with the SonicMoE and ScatterMoE backends, standard training with global-norm clipping (baseline) vs.\ \sonar{}. (a)~Training loss on OLMoE; (b)~peak GPU memory; (c)~throughput averaged over steps 201--2{,}000.}
  \label{fig:pipeline}
  \end{figure*}

  Mixture-of-Experts (MoE) models scale their parameter count without a matching increase in per-token compute, since each token is routed to only a few of many experts~\cite{shazeer2017outrageously,lepikhin2021gshard,fedus2022switch,jiang2024mixtral,deepseekai2024v3}. The cost is paid in memory. A model that activates one billion parameters per token still has to keep the weights, gradients and optimizer state of every parameter during training. In OLMoE-1B-7B~\citep{muennighoff2024olmoe}, for example, the experts hold 6.44B of the 6.92B parameters, and their bf16 gradients alone take 12.9~GB. On a single GPU, this gradient memory alone can decide whether fine-tuning fits at all.

  Much of this gradient memory is held for no reason. A standard training step runs the whole backward pass and only then calls the optimizer, so the gradient of the last MoE layer waits in memory until the backward pass has reached the first layer. An element-wise optimizer such as AdamW~\citep{loshchilov2019adamw} does not need this wait, since the update of a tensor depends only on that tensor's gradient and optimizer state. Prior work on dense models fuses the update into the backward pass~\citep{lv2024lomo,lv2024adalomo,liao2023memory}, and PyTorch provides a hook that fires as soon as a parameter's gradient is ready~\citep{paszke2019pytorch}. Updating every parameter this way frees all gradient memory, but it gives up global-norm clipping, because no gradient survives to the end of the step. In an MoE the experts hold almost all of the gradient memory, so it is enough to update only the experts early.

\begin{figure*}[t]
  \centering
  \begin{tikzpicture}
  \node[anchor=south west, inner sep=0] (img) at (0,0) {\includegraphics[width=0.92\linewidth,trim={0 0 509bp 0},clip]{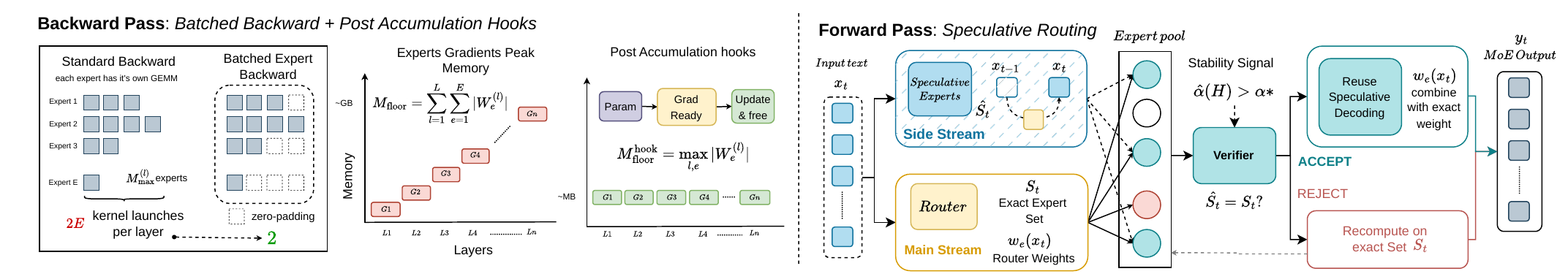}};
  \begin{scope}[x={(img.south east)}, y={(img.north west)}]
  \fill[white] (0.768,0.350) rectangle (0.957,0.510);
  \node[font=\scriptsize] at (0.861, 0.425) {$\displaystyle M_\text{floor}^\text{hook} \le \max_{l}\sum_{e=1}^{E}\lvert W^{(l)}_e\rvert$};
  \end{scope}
  \end{tikzpicture}
  \caption{The \sonar{} backward pass. \emph{Left:} a per-expert layout launches $2E$ weight-gradient kernels per layer, while the batched layout computes all experts with one grouped GEMM per weight tensor. \emph{Middle:} in standard training, expert gradients pile up layer by layer until the optimizer step, reaching $M_\text{floor}$. \emph{Right:} a post-accumulate hook updates each batched tensor and frees its gradient as soon as it is ready, so retained gradient memory stays at the scale of one layer.}
  \label{fig:method}
  \end{figure*}

  We introduce \sonar{}, which updates the experts inside the backward pass. It attaches a hook to each batched expert tensor of every MoE layer, and when the tensor's gradient is ready, the hook applies a fused AdamW update and frees the gradient, so the GPU holds at most one layer of expert gradients at any time (Figure~\ref{fig:method}). All other parameters keep the standard path with global-norm clipping. The placement of the hooks also matters. Expert weights can be stored per expert or as batched tensors~\citep{gale2023megablocks,tan2024scattermoe,guo2025sonicmoe}, and \sonar{} hooks the batched tensors. With $E$ experts per layer and $L$ MoE layers, per-expert hooks fire $2EL$ times per step, while one hook per batched weight tensor fires only $2L$ times. On OLMoE ($E{=}64$, $L{=}16$) this is 32 hook calls per step instead of 2{,}048, for the same update and the same memory.

  In bf16 fine-tuning, \sonar{} lowers peak memory from 55.6 to 43.6~GB on OLMoE and by 21.4\% on Granite-3.1-3B-A800M (Figure~\ref{fig:pipeline}). Its updates are bitwise identical to those of standard training without expert-gradient clipping, and held-out loss stays within 0.003 of every baseline. Each step is also 19--23\% shorter than in standard training with global-norm clipping (Table~\ref{tab:headline}). Two effects shorten the step. The expert gradients skip the global-norm clipping kernels, which gives most of the saving, and on Granite the in-backward updates fill GPU time that would otherwise sit idle, which cuts a further 11.6\% at the same update rule. Applying optimizer-in-backward to all parameters reaches the same memory and speed but cannot clip any gradient. \sonar{} matches it on both while keeping global-norm clipping for the dense parameters. Our contributions are:
  \begin{itemize}
    \item \sonar{}, which updates each batched expert tensor inside the backward pass and frees its gradient at once. Retained expert-gradient memory is bounded by one layer, peak memory falls by 21.6\% (OLMoE) and 21.4\% (Granite), and the updates are bitwise identical to standard training without expert-gradient clipping (\S\ref{sec:method}, \S\ref{sec:main}).
    \item A speedup over standard training, with steps 19--23\% shorter than under global-norm clipping, and a kernel-level profile that explains it. The expert gradients skip the clipping kernels, and on Granite the AdamW kernels run in otherwise idle GPU time (\S\ref{sec:ablation}). Hooking batched tensors rather than single experts keeps this speed, since per-expert hooks make steps 20--72\% slower (\S\ref{sec:granularity}).
    \item A direct comparison with all-parameter optimizer-in-backward, which \sonar{} matches in memory and speed while keeping global-norm clipping, and a measured operating range across batch sizes, gradient checkpointing, 8-bit AdamW and gradient accumulation (\S\ref{sec:composition}, \S\ref{sec:accum}).
  \end{itemize}

\section{Related Work}

\paragraph{MoE training systems.} Systems work on MoE training has made the expert computation faster through better kernels~\citep{gale2023megablocks,tan2024scattermoe,guo2025sonicmoe}, better parallelism and communication across devices~\citep{rajbhandari2022deepspeedmoe,hwang2023tutel,he2022fastermoe,jiang2024lancet,shen2025comet}, and a cheaper expert backward pass~\citep{liu2023sparsemixer,moeblaze2025}. Data-parallel training overlaps the gradient all-reduce with the backward pass~\citep{li2020pytorch}. None of this changes when gradients are released.

\paragraph{Training memory.} ZeRO partitions optimizer state, gradients and parameters across ranks~\citep{rajbhandari2020zero}, and ZeRO-Offload moves them to CPU memory~\citep{ren2021zero}. Activation checkpointing and reversible blocks reduce activation memory~\citep{chen2016sublinear,liu2025revffn}, 8-bit optimizers shrink optimizer state~\citep{dettmers2022optimizers}, and Collage trains fully in bf16 without fp32 master weights~\citep{yu2024collage}. Other methods train fewer parameters or a smaller subspace: LoRA and QLoRA train low-rank adapters~\citep{hu2021lora,dettmers2023qlora}, ESFT fine-tunes only the experts most relevant to a task~\citep{wang2024let}, GaLore projects gradients to a low-rank subspace~\citep{zhao2024galore}, and BAdam updates one block at a time~\citep{luo2024badammemoryefficientparameter}. \sonar{} updates every expert in full and changes only when its gradient is freed, so these savings add to ours (\S\ref{sec:composition}).

\paragraph{Optimizer updates in the backward pass.} The closest work fuses the optimizer update into the backward pass of dense models. LOMO does this with SGD~\citep{lv2024lomo}, AdaLomo adds adaptive learning rates~\citep{lv2024adalomo}, \citet{liao2023memory} run the optimizer step during back-propagation, and GaLore uses per-layer updates to avoid storing weight gradients~\citep{zhao2024galore}. FORGE fuses gradient computation with optimizer updates in registers~\citep{kukreja2026forge}, whereas \sonar{} uses hooks to update and free materialized gradients of batched expert tensors. PyTorch's post-accumulate hooks make this available for every parameter~\citep{paszke2019pytorch}, which we call all-parameter optimizer-in-backward. We do not claim the in-backward update itself, only the choice of where to apply it. Hooking every parameter frees all gradient memory but rules out global-norm clipping~\citep{pascanu2013difficultytrainingrecurrentneural}, while hooking only the batched expert tensors frees almost as much and keeps clipping for the rest of the model.

  \section{Background}
  \label{sec:background}

    \paragraph{MoE layers.} Let $d$ be the hidden size, $n$ the expert intermediate size, $E$ the number of experts, $K$ the number of experts used per token and $L$ the number of MoE layers. A router scores each token $x_t$, keeps the top-$K$ experts $S_t$ and mixes their outputs with weights $w_e(x_t)$ from the router softmax:
  \begin{equation}
  y_t = \sum_{e \in S_t} w_e(x_t)\, \mathrm{Expert}_e(x_t).
  \end{equation}
  Each expert is a gated MLP with a fused gate/up projection of size $2n \times d$ and a down projection of size $d \times n$. Existing MoE implementations store the experts of a layer in two batched tensors, $W^{(l)}_\text{up} \in \mathbb{R}^{E\times 2n\times d}$ and $W^{(l)}_\text{down} \in \mathbb{R}^{E\times d\times n}$, and compute each weight gradient with one grouped GEMM over all experts instead of $E$ separate GEMMs~\citep{gale2023megablocks,tan2024scattermoe,guo2025sonicmoe}. \sonar{} keeps this layout and attaches its hooks to these two tensors.

  \paragraph{Memory in a training step.} A training step keeps four kinds of data on the GPU. Parameters and optimizer state stay resident across steps. Activations are freed during backward as soon as they have been used. Gradients stay in memory until the optimizer step, so by the end of backward every gradient of the model is held at once. In our setting, weights, gradients and AdamW states are all kept in bf16, so the gradients take as much memory as the weights. This gradient memory is what \sonar{} releases early.

  \section{Method}
  \label{sec:method}

  \subsection{Gradient Memory in Standard Training}

  Write $\lvert W^{(l)}_e\rvert$ for the number of weights of expert $e$ in layer $l$. A standard step keeps every expert gradient until the optimizer runs, so at the end of backward the GPU holds
  \begin{equation}
  \label{eq:M_floor}
  M_\text{floor} = \sum_{l=1}^{L}\sum_{e=1}^{E}\lvert W^{(l)}_e\rvert
  \end{equation}
  gradient elements for the experts alone. For OLMoE ($L{=}16$, $E{=}64$, $d{=}2048$, $n{=}1024$) this is 6.44B elements, 12.9~GB in bf16. None of these gradients is needed after its own update, and for an element-wise optimizer that update could have happened the moment the gradient was complete.

  \subsection{In-Backward Expert Updates}

  \sonar{} removes the floor with PyTorch's \texttt{register\_post\_accumulate\_grad\_hook}, which calls a function as soon as a parameter's gradient has been fully accumulated. We register one hook on each batched expert tensor, $W^{(l)}_\text{up}$ and $W^{(l)}_\text{down}$, and remove these tensors from the end-of-step optimizer. When the hook fires, it applies a fused AdamW update to the tensor using its gradient and optimizer state, and then sets the gradient to \texttt{None} so its memory can be reused. All other parameters, such as attention, embeddings and routers, keep the usual end-of-step update. Algorithm~\ref{alg:bb} summarizes the training step.

  Because the backward pass visits the MoE layers one at a time, each expert gradient is freed before the next layer's gradients are computed. The retained expert-gradient memory is therefore bounded by one layer:
  \begin{equation}
  \label{eq:M_hook}
  M_\text{floor}^\text{hook} \le \max_{l}\sum_{e=1}^{E}\lvert W^{(l)}_e\rvert \approx \frac{M_\text{floor}}{L}.
  \end{equation}
  For OLMoE this is a 16-fold reduction, from 12.9~GB to 0.8~GB (one layer). Without gradient accumulation, this is the same update that standard training applies when its clipping excludes the expert tensors, and we verify that the two are bitwise identical (Table~\ref{tab:exactness}). Because the hooks read the learning rate during the backward pass, the learning rate for step $t$ must be set before backward rather than after it.

  \begin{algorithm}[t]
  \small
  \caption{\sonar{} training step.}
  \label{alg:bb}
  \begin{algorithmic}[1]
  \State \textbf{Setup:} for each MoE layer $l$, register a hook on $W^{(l)}_{\text{up}}$ and $W^{(l)}_{\text{down}}$ ($2L$ hooks); exclude them from the end-of-step optimizer.
  \State \textbf{Hook}$(W)$: fused AdamW update of $W$ with $W.\text{grad}$; set $W.\text{grad} \gets \textsc{None}$.
  \State Set the learning rate $\eta_t$ of the hooked and end-of-step optimizers.
  \State Forward pass as usual.
  \For{each MoE layer $l = L,\dots,1$ in backward}
    \State $\nabla W^{(l)}_{\text{down}} \gets$ one grouped GEMM over all experts
    \State \textbf{Hook}$(W^{(l)}_{\text{down}})$ \Comment{update, then free}
    \State $\nabla W^{(l)}_{\text{up}} \gets$ one grouped GEMM over all experts
    \State \textbf{Hook}$(W^{(l)}_{\text{up}})$ \Comment{update, then free}
    \State pass the input gradient to layer $l-1$
  \EndFor
  \State Optimizer step on non-expert parameters only.
  \end{algorithmic}
  \end{algorithm}

  \subsection{Hook Granularity}
  \label{sec:granularity}

  The same schedule can be built on per-expert parameters, with one hook and one optimizer per expert slice, so each step makes $2EL$ optimizer calls instead of $2L$. Since AdamW is element-wise, one fused update on an $(E, 2n, d)$ tensor gives exactly the same result as $E$ fused updates on its slices. What changes is the number of hook calls per step:
  \begin{equation}
  \label{eq:hook_count}
  N_\text{hooks}^\text{per-expert} = 2EL, \qquad N_\text{hooks}^\text{batched}=2L.
  \end{equation}
  Every call has a fixed cost, because autograd has to schedule a Python callback and the callback launches its own small kernels. If $\tau_\text{hook}$ is this fixed cost, the per-expert layout pays $2EL\,\tau_\text{hook}$ per step and the batched layout pays $2L\,\tau_\text{hook}$. For OLMoE this means 2{,}048 calls instead of 32, each doing $E$ times less work. Modern MoE designs favour many small experts~\citep{dai2024deepseekmoe,deepseekai2024v3}, which makes the gap grow with $E$. On OLMoE ($E{=}64$), per-expert hooks add 23.6\,ms per step (+20\%), about 12\,$\mu$s per extra optimizer call. On Granite ($E{=}40$, 32 layers) they add 61.1\,ms (+72\%), about 24\,$\mu$s per extra call. Both layouts use the same memory and the per-expert layout is slower (Table~\ref{tab:granularity}), so \sonar{} attaches the hooks to the batched tensors.

  \subsection{Clipping and Gradient Accumulation}
  \label{sec:changes}

  Updating during backward changes two things that training loops usually take for granted, and our evaluation keeps both visible.

  \paragraph{Global-norm clipping.} Clipping by the global gradient norm needs every gradient before any update, but \sonar{} has already updated and freed the expert gradients by then. We therefore compare with two standard baselines, one that clips all gradients and one that clips only the non-expert gradients. The second performs exactly our update (\S\ref{sec:setup}).

  \paragraph{Gradient accumulation.} Gradient accumulation sums the gradients of $A$ microbatches and applies one optimizer step to their mean $\bar g = \tfrac{1}{A}\sum_a g_a$. A post-accumulate hook fires as soon as each microbatch's gradient is ready, so the hook schedule applies one AdamW update per microbatch instead. These are not equivalent to a single update using the combined gradients, because both the weights and AdamW's optimizer state change between microbatches. The first and second moments can both differ between these schedules. The standard schedule updates $v_t$ once per step with $\bar g^2$, while the hook schedule updates it $A$ times with the individual $g_a^2$:
  \begin{equation}
  \label{eq:gradacc}
  \begin{aligned}
  v_t^\text{std} &= \beta_2\, v_{t-1} + (1-\beta_2)\,\bar g^{2},\\
  v_t^\text{hook} &= \beta_2^{A}\, v_{t-1} + (1-\beta_2)\sum_{a=1}^{A}\beta_2^{A-a}\,g_a^{2}.
  \end{aligned}
  \end{equation}
  For independent microbatch gradient noise of equal variance, averaging $A$ microbatches divides the noise variance in $\bar g$ by $A$, whereas the hook schedule uses the individual noisy gradients in its second-moment updates. The two schedules coincide at $A{=}1$. We therefore use $A{=}1$ in all main experiments and evaluate $A{>}1$ separately in \S\ref{sec:accum}. Table~\ref{tab:accum} gives the measured cost of these per-microbatch updates.

  \section{Experiments}
  \label{sec:experiments}

  \subsection{Experimental Setup}
  \label{sec:setup}

  \paragraph{Models and data.} We pick two public MoE models of different sizes and expert counts that both run on batched expert kernels. We fine-tune \nolinkurl{allenai/OLMoE-1B-7B-0924}~\citep{muennighoff2024olmoe} (16 MoE layers, 64 experts, top-8) and \nolinkurl{ibm-granite/granite-3.1-3b-a800m-base}~\citep{granite2024} (32 MoE layers, 40 experts) on 8{,}000 Alpaca examples~\citep{taori2023stanford} for 2{,}000 steps. A held-out set of 500 further examples (indices 8000--8499 of the same seed-42 shuffle) is disjoint from training. We use a batch size of 2, sequences of at most 256 tokens and no gradient accumulation. The learning rate peaks at $2\times10^{-5}$ after 100 warm-up steps and decays with a cosine schedule toward $2\times10^{-6}$. AdamW uses $\beta{=}(0.9, 0.95)$, $\epsilon{=}10^{-8}$ and no weight decay. Training is in bf16, with weights, gradients and AdamW states all kept in bf16 and no fp32 master weights. All arms use seed 42 and the same data order, which we check by hashing the first batches.

  \paragraph{Baselines.} \emph{Standard training with global clipping} runs the backward pass, clips the global gradient norm of all parameters to 1.0 and takes one fused AdamW step. This is the usual default. \emph{Standard training without expert-gradient clipping} is the same but clips only the non-expert parameters. It performs exactly the \sonar{} update, so any speed difference between the two comes from when the update happens rather than from the update itself. \emph{All-parameter optimizer-in-backward} updates every parameter in its own backward hook, so no clipping is possible. We also run \sonar{} with \emph{per-expert hooks}, one optimizer per expert slice, as an ablation.

  \paragraph{Backends.} We run every method with two expert-kernel libraries that store experts in batched tensors, SonicMoE~\citep{guo2025sonicmoe} and ScatterMoE~\citep{tan2024scattermoe}. Table~\ref{tab:headline} reports both; the other experiments use SonicMoE.

  \paragraph{Metrics and hardware.} We report the median step time over steps 201--2{,}000, with a GPU synchronization at the end of each step, throughput in supervised response tokens per second (about 111 per step), and peak allocated and peak reserved GPU memory. Same-seed repeats vary by at most 1.4\% in step time and 0.0016 in held-out loss. Runs use two NVIDIA RTX PRO 6000 Blackwell GPUs (96~GB) with PyTorch 2.11 and CUDA 13.1. All arms of one model and backend ran on the same GPU, and the two GPUs have different power limits, so we compare step times only within a model and backend. Downstream accuracy is measured with lm-evaluation-harness~\citep{gao2021framework} on the full test sets of ARC-Challenge~\citep{clark2018think} and HellaSwag~\citep{zellers2019hellaswag} (0-shot, \texttt{acc\_norm}) and MMLU~\citep{hendrycks2020measuring} (5-shot, accuracy).

  \subsection{Main Results}
  \label{sec:main}

  \begin{table*}[t]
  \centering
  \small
  \setlength{\tabcolsep}{3.5pt}
  \caption{Alpaca fine-tuning for 2{,}000 steps with batch size 2 and seed 42 on two expert-kernel backends; every row uses the same data order. $\Delta$ is relative to standard training with global clipping on the same model and backend. Step time is the median over steps 201--2{,}000. Same-seed repeats vary by at most 1.4\% in step time and 0.0016 in held-out loss. All rows of one model and backend ran on one GPU; the OLMoE SonicMoE rows ran on a different GPU from all other rows, so compare step times only within a model and backend. Bold marks the lowest memory and step time per model and backend.}
  \label{tab:headline}
  \resizebox{\textwidth}{!}{%
  \begin{tabular}{lllccccc}
  \toprule
  Model & Backend & Method & Peak alloc.\ GB ($\Delta$) & Peak res.\ GB & Step ms ($\Delta$) & Tok/s & Held-out loss \\
  \midrule
  \multirow{8}{*}{OLMoE-1B-7B} & \multirow{4}{*}{SonicMoE} & Standard (global clip) & 55.55 & 56.42 & 142.8 & 768 & 1.2577 \\
  & & Standard (no expert clip) & 55.55 (0.0\%) & 56.42 & 115.9 ($-$18.8\%) & 958 & 1.2591 \\
  & & Optimizer-in-backward (all) & \textbf{43.35 ($-$22.0\%)} & 44.24 & \textbf{114.4 ($-$19.9\%)} & 970 & 1.2612 \\
  & & \sonar{} & 43.57 ($-$21.6\%) & 44.24 & 116.1 ($-$18.7\%) & 948 & 1.2596 \\
  \cmidrule(l){2-8}
  & \multirow{4}{*}{ScatterMoE} & Standard (global clip) & 55.55 & 57.17 & 160.2 & 688 & 1.2585 \\
  & & Standard (no expert clip) & 55.55 (0.0\%) & 57.17 & 133.3 ($-$16.7\%) & 825 & 1.2601 \\
  & & Optimizer-in-backward (all) & \textbf{43.60 ($-$21.5\%)} & 44.79 & \textbf{131.0 ($-$18.2\%)} & 841 & 1.2610 \\
  & & \sonar{} & 43.81 ($-$21.1\%) & 44.80 & 133.2 ($-$16.8\%) & 826 & 1.2601 \\
  \midrule
  \multirow{8}{*}{Granite-3.1-3B} & \multirow{4}{*}{SonicMoE} & Standard (global clip) & 26.91 & 28.03 & 109.1 & 1111 & 1.1625 \\
  & & Standard (no expert clip) & 26.91 (0.0\%) & 28.02 & 95.5 ($-$12.4\%) & 1275 & 1.1643 \\
  & & Optimizer-in-backward (all) & 21.15 ($-$21.4\%) & 21.90 & 86.5 ($-$20.7\%) & 1421 & 1.1630 \\
  & & \sonar{} & \textbf{21.14 ($-$21.4\%)} & 22.03 & \textbf{84.4 ($-$22.7\%)} & 1450 & 1.1616 \\
  \cmidrule(l){2-8}
  & \multirow{4}{*}{ScatterMoE} & Standard (global clip) & 26.89 & 28.49 & 106.8 & 1151 & 1.1591 \\
  & & Standard (no expert clip) & 26.89 (0.0\%) & 28.49 & 93.7 ($-$12.3\%) & 1230 & 1.1582 \\
  & & Optimizer-in-backward (all) & 21.77 ($-$19.1\%) & 22.40 & 90.0 ($-$15.8\%) & 1362 & 1.1585 \\
  & & \sonar{} & \textbf{21.75 ($-$19.1\%)} & 22.34 & \textbf{88.2 ($-$17.4\%)} & 1390 & 1.1582 \\
  \bottomrule
  \end{tabular}}
  \end{table*}

  \sonar{} saves the memory that the mechanism predicts (Table~\ref{tab:headline}). With SonicMoE, peak memory falls by 12.0~GB on OLMoE and 5.8~GB on Granite, which is the expert-gradient buffer of 12.9 and 6.0~GB minus the one layer that is still held. Held-out loss stays within 0.003 of every baseline. Speed needs a more careful reading. Standard training without expert-gradient clipping is as fast as \sonar{} on OLMoE (115.9 vs.\ 116.1~ms), so on OLMoE the whole speedup over standard training comes from not clipping the expert gradients. On Granite \sonar{} is a further 11.6\% faster than that baseline, which \S\ref{sec:ablation} traces to idle GPU time. All-parameter optimizer-in-backward matches \sonar{} in both memory and speed and differs only in giving up global-norm clipping on the dense parameters. ScatterMoE shows the same pattern. \sonar{} lowers peak memory by 21.1\% on OLMoE and 19.1\% on Granite, is as fast as standard training without expert-gradient clipping on OLMoE (133.2 vs.\ 133.3~ms), and is 5.9\% faster than that baseline on Granite (88.2 vs.\ 93.7~ms), a smaller gain than with SonicMoE.

  \begin{table}[t]
  \centering
  \small
  \setlength{\tabcolsep}{3pt}
  \caption{Downstream accuracy on OLMoE after 2{,}000 steps (full test sets). Standard errors are 0.015 for ARC-Challenge and 0.004 for HellaSwag and MMLU in every row.}
  \label{tab:downstream}
  \begin{tabular*}{\columnwidth}{@{\extracolsep{\fill}}lccc@{}}
  \toprule
  Checkpoint & ARC-C & HellaSwag & MMLU \\
  \midrule
  Pretrained & 0.482 & 0.771 & 0.527 \\
  Standard (global clip) & 0.532 & 0.771 & 0.521 \\
  \sonar{} & 0.516 & 0.772 & 0.521 \\
  \bottomrule
  \end{tabular*}
  \end{table}

  Loss alone does not show whether a fine-tuned model is still useful, so we also evaluate the OLMoE checkpoints after 2{,}000 steps on downstream tasks (Table~\ref{tab:downstream}). Fine-tuning raises ARC-Challenge accuracy by 3--5 points and leaves HellaSwag and MMLU almost unchanged. The two training methods differ by 0.016 on ARC-Challenge, about one standard error, and by at most 0.001 on HellaSwag and MMLU. We evaluate only OLMoE and omit GSM8K, where this model stays near 0.1 accuracy after Alpaca fine-tuning.

  \subsection{Sources of the Speedup}
  \label{sec:ablation}

  \begin{figure}[t]
  \centering
  \includegraphics[width=\columnwidth]{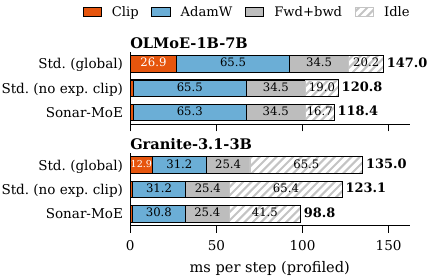}
  \caption{Where step time goes (ms per step): \texttt{torch.profiler} over steps 300--309. Profiling adds CPU overhead, so absolute times are higher than in Table~\ref{tab:headline}; compare bars only. ``Clip'' covers the norm and scaling kernels of global-norm clipping, which take 1.8--1.9~ms on OLMoE and 1.1--1.2~ms on Granite when the expert gradients are not clipped. ``Fwd+bwd'' covers the forward and backward kernels.}
  \label{fig:profile}
  \end{figure}

  \begin{table}[t]
  \centering
  \small
  \setlength{\tabcolsep}{3pt}
  \caption{Hook granularity. Both layouts hold the same memory; per-expert hooks make $2EL$ optimizer calls per step instead of $2L$.}
  \label{tab:granularity}
  \begin{tabular*}{\columnwidth}{@{\extracolsep{\fill}}llrrrr@{}}
  \toprule
  Model & Hooks & Calls & Peak GB & Step ms & Held-out \\
  \midrule
  OLMoE & batched & 32 & 43.57 & 116.1 & 1.2596 \\
  OLMoE & per-expert & 2048 & 43.58 & 139.7 & 1.2600 \\
  Granite & batched & 64 & 21.14 & 84.4 & 1.1616 \\
  Granite & per-expert & 2560 & 21.37 & 145.5 & 1.1606 \\
  \bottomrule
  \end{tabular*}
  \end{table}

  Three effects explain the step times (Figure~\ref{fig:profile}). First, global-norm clipping of the expert gradients costs about 25~ms of kernel time per step on OLMoE (26.9 vs.\ 1.8~ms). It reads the 12.9~GB of gradients to compute the norm, then reads and writes them again to rescale them. This accounts for nearly all of the difference between standard training and \sonar{} on OLMoE. The cost scales with the number of expert parameters rather than with the batch, which is why it matters most for small batches (\S\ref{sec:composition}). Second, on Granite the GPU work is the same with and without in-backward updates (57.7 vs.\ 57.4~ms busy), but idle time falls from 65 to 42~ms. Granite's backward pass is limited by kernel-launch overhead, and the in-backward AdamW kernels run in these gaps. OLMoE leaves little idle time, so it gains almost nothing from this. Third, attaching hooks per expert keeps the memory saving but adds 23.6~ms per step on OLMoE, about 12\,$\mu$s per extra optimizer call (Table~\ref{tab:granularity}).

  \subsection{Exactness and Training Quality}
  \label{sec:dynamics}

  \paragraph{Update exactness.} Without gradient accumulation, \sonar{} performs the same arithmetic as standard training that clips only the non-expert gradients. We train OLMoE for three steps with a learning rate of $10^{-4}$, large enough that bf16 updates are visible, and compare the weights of two runs element by element (Table~\ref{tab:exactness}). \sonar{} and standard training without expert-gradient clipping give exactly the same 6.44B expert weights. They differ in 6 of the 477M other weights, and two runs of the same baseline also differ in 6 elements, so this is ordinary run-to-run noise, not an effect of the schedule. Per-expert and batched hooks also give the same expert weights, and all-parameter optimizer-in-backward gives exactly the same weights as standard training without clipping.

  \begin{table}[h]
  \centering
  \small
  \setlength{\tabcolsep}{3pt}
  \caption{Weights that differ between two runs after three steps on OLMoE (6.44B expert weights, 477M other weights).}
  \label{tab:exactness}
  \resizebox{\columnwidth}{!}{%
  \begin{tabular}{lrr}
  \toprule
  Runs compared & Expert weights & Other weights \\
  \midrule
  Same baseline, run twice & 0 & 6 \\
  \sonar{} vs.\ standard (no expert clip) & 0 & 6 \\
  \sonar{}: per-expert vs.\ batched hooks & 0 & 5 \\
  Opt.-in-backward (all) vs.\ standard (no clip) & 0 & 0 \\
  Standard: global clip vs.\ no expert clip & 3.04B (47\%) & 91.5M (19\%) \\
  \bottomrule
  \end{tabular}}
  \end{table}

  Clipping the expert gradients, as standard training does, changes 47\% of the expert weights after three steps, yet changes held-out loss on OLMoE by only 0.0014 after 2{,}000 steps (Table~\ref{tab:headline}). AdamW divides each update by a running estimate of the gradient scale, which largely cancels a uniform rescaling.

  \paragraph{Pure-bf16 updates.} Because weights are stored in bf16, an update smaller than half a unit in the last place of a weight is rounded away. Table~\ref{tab:bf16} shows that after 2{,}000 steps most weights still hold their pretrained values, and that the two methods change the same share of weights.

  \begin{table}[h]
  \centering
  \small
  \caption{Share of weights that differ from their pretrained values after 2{,}000 steps of bf16 training on OLMoE.}
  \label{tab:bf16}
  \begin{tabular}{lcc}
  \toprule
  Method & Expert weights & Other weights \\
  \midrule
  Standard (global clip) & 19.3\% & 22.4\% \\
  \sonar{} & 19.3\% & 22.5\% \\
  \bottomrule
  \end{tabular}
  \end{table}

  \subsection{Batch Size and Memory Composition}
  \label{sec:composition}

  \begin{figure}[t]
  \centering
  \includegraphics[width=\columnwidth]{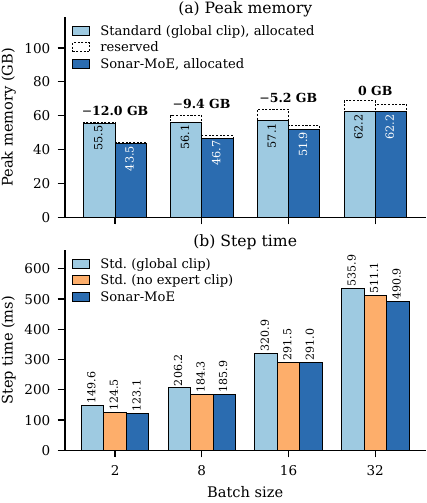}
  \caption{Batch size on OLMoE. (a)~Peak allocated memory (filled bars) and peak reserved memory (dashed outlines) for standard training and \sonar{}. The bold labels give the saving in allocated memory. (b)~Step time in ms for each method. Sequences are padded to 256 tokens (512--8{,}192 tokens per step) so that kernel autotuning happens once, in warm-up.}
  \label{fig:batch}
  \end{figure}

  \begin{table}[t]
  \centering
  \small
  \setlength{\tabcolsep}{3pt}
  \caption{Memory composition on OLMoE at batch size 2 (300-step runs). Peak allocated and reserved memory in GB, step time in ms, and whether 300 steps completed with the PyTorch allocator capped at 46~GB (standing for a 48~GB GPU) or 30.5~GB (a 32~GB GPU). GC = gradient checkpointing; opt.-in-bwd = all-parameter optimizer-in-backward.}
  \label{tab:memory}
  \resizebox{\columnwidth}{!}{%
  \begin{tabular}{lrrrcc}
  \toprule
  & & & & \multicolumn{2}{c}{Runs under cap} \\
  \cmidrule(lr){5-6}
  Setup & Alloc. & Res. & ms & 46~GB & 30.5~GB \\
  \midrule
  Standard & 55.55 & 56.40 & 140.8 & OOM & -- \\
  Standard + GC & 55.48 & 56.16 & 155.6 & -- & -- \\
  \sonar{} & 43.56 & 44.22 & 111.8 & \checkmark & -- \\
  \sonar{} + GC & 43.23 & 44.03 & 130.1 & -- & -- \\
  Opt.-in-bwd & 43.35 & 44.24 & 109.9 & \checkmark & -- \\
  \midrule
  8-bit AdamW & 41.93 & 43.13 & 186.0 & -- & OOM \\
  8-bit + \sonar{} & 29.94 & 30.52 & 165.4 & -- & \checkmark \\
  8-bit + \sonar{} + GC & 29.61 & 29.78 & 189.5 & -- & \checkmark \\
  8-bit + opt.-in-bwd & 29.74 & 30.69 & 165.0 & -- & OOM \\
  \bottomrule
  \end{tabular}}
  \end{table}

  The cost of clipping expert gradients is fixed per step (22--29~ms), so the speedup over standard training shrinks from 18\% at batch size 2 to 8\% at batch size 32 (Figure~\ref{fig:batch}). With the same update rule, \sonar{} and standard training are within $\pm$1.1\% at batch sizes 2--16, and at batch size 32 \sonar{} is 4.0\% faster. The memory saving also shrinks with batch size, because activations, not gradients, set the peak at large batches. The saving in peak allocated memory is 12.0~GB at batch size 2, 9.4~GB at 8, 5.2~GB at 16 and none at 32, where only 2.2~GB of reserved memory is saved. \sonar{} helps most when batches are small relative to the model. In practice the saving can be read off Eq.~\ref{eq:M_floor}: it equals the expert-gradient buffer as long as gradients, not activations, set the peak.

  Gradient checkpointing does not lower peak memory at batch size 2, because activations are small, and only costs time (+10--16\%, Table~\ref{tab:memory}). 8-bit AdamW and \sonar{} save different memory, optimizer states and gradients, so they combine. Together they need about 30~GB and run under a 30.5~GB cap, where 8-bit AdamW alone runs out of memory. All-parameter optimizer-in-backward gives the same savings. It also runs under the 46~GB cap, and with 8-bit AdamW its peak allocated memory matches ours (29.7 vs.\ 29.9~GB). 8-bit AdamW is slower per step (186 vs.\ 141~ms).

  \paragraph{8-bit AdamW quality.} For the combination that fits OLMoE fine-tuning on a 32~GB GPU, we also check that quality is preserved. We use the setup of Table~\ref{tab:headline} (2{,}000 steps, batch size 2, seed 42) but store the AdamW states in 8 bits with bitsandbytes \texttt{AdamW8bit}~\citep{dettmers2022optimizers}. With \sonar{}, each expert tensor gets its own 8-bit optimizer inside its hook, and the other parameters use one 8-bit optimizer at the end of the step with global-norm clipping, as in Table~\ref{tab:headline}. Held-out loss stays within 0.003 of standard training (Table~\ref{tab:8bit}), so the memory saving of 8-bit AdamW and \sonar{} comes at no visible cost in quality.

  \begin{table}[h]
  \centering
  \small
  \caption{Held-out loss on OLMoE after 2{,}000 steps with 8-bit AdamW, compared with standard training from Table~\ref{tab:headline}.}
  \label{tab:8bit}
  \begin{tabular}{lc}
  \toprule
  Configuration & Held-out loss \\
  \midrule
  Standard (global clip) & 1.2577 \\
  8-bit AdamW & 1.2574 \\
  8-bit AdamW + \sonar{} & 1.2601 \\
  \bottomrule
  \end{tabular}
  \end{table}

  \subsection{Gradient Accumulation}
  \label{sec:accum}

  \begin{table}[t]
  \centering
  \small
  \setlength{\tabcolsep}{2pt}
  \caption{Gradient accumulation with $A$ microbatches per step (OLMoE, 500 steps). Step time in ms; peak allocated memory in GB for standard training (both variants) / \sonar{}. \sonar{} makes $A$ updates per step.}
  \label{tab:accum}
  \vskip 0.15in
  \begin{tabular}{rrrrc}
  \toprule
  $A$ & Std.\ (global) & Std.\ (no exp.) & \sonar{} & Peak GB \\
  \midrule
  1 & 143.4 & 117.3 & 116.4 & 55.55 / 43.57 \\
  2 & 217.8 & 191.9 & 229.4 & 57.21 / 44.31 \\
  4 & 368.5 & 341.7 & 451.7 & 57.21 / 44.31 \\
  8 & 665.8 & 639.6 & 895.6 & 57.21 / 44.32 \\
  \bottomrule
  \end{tabular}
  \end{table}

  As described in \S\ref{sec:changes}, hooks fire once per microbatch, so with $A$ microbatches \sonar{} takes $A$ optimizer steps where the baseline takes one. Table~\ref{tab:accum} sweeps $A \in \{1,2,4,8\}$. At $A{=}1$ \sonar{} is as fast as standard training without expert-gradient clipping (116.4 vs.\ 117.3~ms). With accumulation it runs the optimizer $A$ times per step and is slower than standard training with global clipping, by 5\% at $A{=}2$, 23\% at $A{=}4$ and 35\% at $A{=}8$, while still saving 12.9~GB. Each extra update is a full AdamW pass over the 6.44B expert weights, about 37~ms at every $A$, so the overhead grows linearly with $A$. The method also becomes a different optimizer, so we recommend \sonar{} for $A{=}1$, or for settings where memory is the binding constraint and the change of update rule is acceptable.

  \section{Discussion}
  \label{sec:discussion}

  \paragraph{Practical memory gains.} The most direct benefit is that a run fits where it did not before. Fine-tuning OLMoE-1B-7B in our setup peaks at 55.6~GB with standard training and at 43.6~GB with \sonar{}, which fits on a 48~GB GPU, since it runs under a 46~GB allocator cap where standard training runs out of memory. With 8-bit AdamW it fits a 32~GB GPU (30.5~GB cap), where 8-bit AdamW alone does not. All-parameter optimizer-in-backward reaches the same peak memory and also fits under the 46~GB cap. Gradient checkpointing does not reduce peak memory at batch size 2 (\S\ref{sec:composition}), and the saving shrinks as batches grow, because activations then set the peak.

  \paragraph{Operating conditions.} Our results point to a simple rule. Early release is a good default when four conditions hold: the optimizer is element-wise, gradients are not accumulated across microbatches, the experts are stored in batched tensors, and the expert gradients are not needed for global-norm clipping. The memory saving is largest at small batch sizes. A speed benefit appears only when the baseline clips expert gradients or when the backward pass leaves the GPU idle. With accumulation the method becomes a different and slower optimizer (\S\ref{sec:accum}), and with per-expert hooks it keeps the memory saving but loses time (\S\ref{sec:ablation}). We think this kind of explicit operating range is more useful to practitioners than a single headline number, and we have tried to report every regime we measured, including the ones where the method does not help.

  \paragraph{Why hook granularity is overlooked.} Optimizer-in-backward methods were developed for dense models, where each weight matrix is one parameter and the question of hook granularity never arises. In MoE code it does arise. Expert weights are often stored as a list of per-expert modules for readability, or are split into per-expert views when a checkpoint is loaded. Both cases silently multiply the hook count by $E$. Our ablation shows that this keeps the memory saving intact but adds 20\% (OLMoE) and 72\% (Granite) step time (Table~\ref{tab:granularity}), so the problem is easy to overlook if only memory is measured.

  \section{Conclusion}

  In MoE training, the gradients of all experts usually sit in memory until the optimizer step, even though each one is needed only for its own update. \sonar{} updates every batched expert tensor as soon as its gradient is ready and frees the gradient right away, which bounds the retained expert-gradient memory by one layer. At batch size 2 it lowers peak memory by 21.6\% on OLMoE and 21.4\% on Granite, and its updates are bitwise identical to standard training without expert-gradient clipping. Its speed advantage over standard training comes from not clipping the expert gradients and, on models whose backward pass leaves the GPU idle, from running the optimizer kernels in that idle time. Hooks attached per expert save the same memory but make each step 20--72\% slower, so early release should be attached to batched expert tensors. All-parameter optimizer-in-backward matches the method in memory and speed. Natural next steps are to support global-norm clipping without keeping all gradients in memory, and to support data parallelism by updating after the gradient all-reduce. We are also extending \sonar{} to inference, where expert computation can be started speculatively before the router's decision is final and reconciled against the exact routing, so that the same early-release principle speeds up both training and inference of MoE models.

  \section*{Limitations}

  \textbf{Gradient clipping.} Standard global-norm clipping needs all gradients before any update, while \sonar{} updates experts during backward. Compatibility can be restored by delaying the update and keeping, offloading or recomputing the gradients after the norm is known, or by clipping each expert tensor by its own norm inside the hook. We plan to evaluate these options in the next version. \textbf{Gradient accumulation.} With $A{\ge}2$ the method applies $A$ updates per step and becomes a different optimizer that is 5--35\% slower than standard training with global clipping (\S\ref{sec:accum}), so our main results use $A{=}1$. The schedule is exact for element-wise optimizers such as AdamW, SGD or Lion, and extending it to optimizers that need statistics across tensors is left to future work. \textbf{Scope of the evidence.} All results use one GPU, and under data parallelism the update would run after the gradient all-reduce, which we will add by integrating with the communication hooks. The memory saving is largest at small batch sizes and shrinks as activations dominate. We train in bf16 and test models up to 7B parameters on Alpaca fine-tuning with one seed, using same-seed repeats to measure noise; larger models, more datasets and more seeds are planned for the next experiments.

  \bibliographystyle{mlsys2025}
  \bibliography{example_paper}

@inproceedings{shazeer2017outrageously,
  title     = {Outrageously Large Neural Networks: The Sparsely-Gated
               Mixture-of-Experts Layer},
  author    = {Shazeer, Noam and Mirhoseini, Azalia and Maziarz, Krzysztof
               and Davis, Andy and Le, Quoc and Hinton, Geoffrey and Dean, Jeff},
  booktitle = {International Conference on Learning Representations},
  year      = {2017},
  url       = {https://arxiv.org/abs/1701.06538}
}

@article{fedus2022switch,
  title     = {Switch Transformers: Scaling to Trillion Parameter Models
               with Simple and Efficient Sparsity},
  author    = {Fedus, William and Zoph, Barret and Shazeer, Noam},
  journal   = {Journal of Machine Learning Research},
  volume    = {23},
  number    = {120},
  pages     = {1--39},
  year      = {2022},
  url       = {https://arxiv.org/abs/2101.03961}
}

@article{jiang2024mixtral,
  title     = {Mixtral of Experts},
  author    = {Jiang, Albert Q. and Sablayrolles, Alexandre and Roux, Antoine
               and Mensch, Arthur and Savary, Blanche and Bamford, Chris
               and Chaplot, Devendra Singh and de las Casas, Diego and
               Bou Hanna, Emma and Bressand, Florian and others},
  journal   = {arXiv preprint arXiv:2401.04088},
  year      = {2024},
  url       = {https://arxiv.org/abs/2401.04088}
}

@article{liu2023sparsemixer,
  title     = {Sparse Backpropagation for {MoE} Training},
  author    = {Liu, Liyuan and Gao, Jianfeng and Chen, Weizhu},
  journal   = {arXiv preprint arXiv:2310.00811},
  year      = {2023},
  url       = {https://arxiv.org/abs/2310.00811}
}

@article{liao2023memory,
  title     = {Memory Efficient Mixed-Precision Optimizers},
  author    = {Lewandowski, Basile and Kosson, Atli},
  journal   = {arXiv preprint arXiv:2309.12381},
  year      = {2023},
  url       = {https://arxiv.org/abs/2309.12381}
}

@article{moeblaze2025,
  title     = {{MoEBlaze}: Breaking the Memory Wall for Efficient {MoE} Training on Modern {GPUs}},
  author    = {Zhang, Jiyuan and Liu, Yining and Yan, Siqi and Deng, Lisen and Cao, Jennifer and Yang, Shuqi and Ni, Min and Xue, Bi and Li, Shen},
  journal   = {arXiv preprint arXiv:2601.05296},
  year      = {2026},
  url       = {https://arxiv.org/abs/2601.05296}
}

@article{guo2025sonicmoe,
  title={{SonicMoE}: Accelerating {MoE} with {IO} and Tile-aware Optimizations},
  author={Guo, Wentao and Mishra, Mayank and Cheng, Xinle and Stoica, Ion and Dao, Tri},
  journal={arXiv preprint arXiv:2512.14080},
  year={2025}
}

@article{tan2024scattermoe,
  title={Scattered mixture-of-experts implementation},
  author={Tan, Shawn and Shen, Yikang and Panda, Rameswar and Courville, Aaron},
  journal={arXiv preprint arXiv:2403.08245},
  year={2024}
}

@inproceedings{gale2023megablocks,
  title     = {{MegaBlocks}: Efficient Sparse Training with
               Mixture-of-Experts},
  author    = {Gale, Trevor and Narayanan, Deepak and Young, Cliff and
               Zaharia, Matei},
  booktitle = {Proceedings of Machine Learning and Systems (MLSys)},
  year      = {2023},
  url       = {https://arxiv.org/abs/2211.15841}
}

@inproceedings{lepikhin2021gshard,
  title     = {{GShard}: Scaling Giant Models with Conditional Computation
               and Automatic Sharding},
  author    = {Lepikhin, Dmitry and Lee, HyoukJoong and Xu, Yuanzhong and
               Chen, Dehao and Firat, Orhan and Huang, Yanping and
               Krikun, Maxim and Shazeer, Noam and Chen, Zhifeng},
  booktitle = {International Conference on Learning Representations},
  year      = {2021},
  url       = {https://arxiv.org/abs/2006.16668}
}

@article{muennighoff2024olmoe,
  title     = {{OLMoE}: Open Mixture-of-Experts Language Models},
  author    = {Muennighoff, Niklas and Soldaini, Luca and Groeneveld, Dirk
               and Lo, Kyle and Morrison, Jacob and Min, Sewon and
               Shi, Weijia and Walsh, Pete and Tafjord, Oyvind and
               Lambert, Nathan and others},
  journal   = {arXiv preprint arXiv:2409.02060},
  year      = {2024},
  url       = {https://arxiv.org/abs/2409.02060}
}

@inproceedings{dai2024deepseekmoe,
  title     = {{DeepSeekMoE}: Towards Ultimate Expert Specialization in
               Mixture-of-Experts Language Models},
  author    = {Dai, Damai and Deng, Chengqi and Zhao, Chenggang and Xu, R. X.
               and Gao, Huazuo and Chen, Deli and Li, Jiashi and others},
  booktitle = {Proceedings of the 62nd Annual Meeting of the Association
               for Computational Linguistics (ACL)},
  year      = {2024},
  url       = {https://arxiv.org/abs/2401.06066}
}

@article{deepseekai2024v3,
  title     = {{DeepSeek-V3} Technical Report},
  author    = {{DeepSeek-AI}},
  journal   = {arXiv preprint arXiv:2412.19437},
  year      = {2024},
  url       = {https://arxiv.org/abs/2412.19437}
}

@inproceedings{rajbhandari2022deepspeedmoe,
  title     = {{DeepSpeed-MoE}: Advancing Mixture-of-Experts Inference and
               Training to Power Next-Generation {AI} Scale},
  author    = {Rajbhandari, Samyam and Li, Conglong and Yao, Zhewei and
               Zhang, Minjia and Aminabadi, Reza Yazdani and Awan, Ammar
               Ahmad and Rasley, Jeff and He, Yuxiong},
  booktitle = {Proceedings of the 39th International Conference on Machine
               Learning (ICML)},
  year      = {2022},
  url       = {https://arxiv.org/abs/2201.05596}
}

@inproceedings{hwang2023tutel,
  title     = {{Tutel}: Adaptive Mixture-of-Experts at Scale},
  author    = {Hwang, Changho and Cui, Wei and Xiong, Yifan and Yang, Ziyue
               and Liu, Ze and Hu, Han and Wang, Zilong and Salas, Rafael
               and Jose, Jithin and Ram, Prabhat and others},
  booktitle = {Proceedings of Machine Learning and Systems (MLSys)},
  year      = {2023},
  url       = {https://arxiv.org/abs/2206.03382}
}

@inproceedings{shen2025comet,
  title     = {{COMET}: Fine-grained Computation-communication Overlapping for Mixture-of-Experts},
  author    = {Zhang, Shulai and Zheng, Ningxin and Lin, Haibin and Jiang, Ziheng and Bao, Wenlei and Jiang, Chengquan and Hou, Qi and Cui, Weihao and Zheng, Size and Chang, Li-Wen and Chen, Quan and Liu, Xin},
  booktitle = {Proceedings of Machine Learning and Systems (MLSys)},
  year      = {2025}
}

@article{jiang2024lancet,
  title={Lancet: Accelerating mixture-of-experts training via whole graph computation-communication overlapping},
  author={Jiang, Chenyu and Tian, Ye and Jia, Zhen and Zheng, Shuai and Wu, Chuan and Wang, Yida},
  journal={Proceedings of Machine Learning and Systems},
  volume={6},
  pages={74--86},
  year={2024}
}

@inproceedings{rajbhandari2020zero,
  title     = {{ZeRO}: Memory Optimizations Toward Training Trillion Parameter
               Models},
  author    = {Rajbhandari, Samyam and Rasley, Jeff and Ruwase, Olatunji and
               He, Yuxiong},
  booktitle = {Proceedings of the International Conference for High
               Performance Computing, Networking, Storage and Analysis (SC)},
  year      = {2020},
  url       = {https://arxiv.org/abs/1910.02054}
}

@article{chen2016sublinear,
  title     = {Training Deep Nets with Sublinear Memory Cost},
  author    = {Chen, Tianqi and Xu, Bing and Zhang, Chiyuan and
               Guestrin, Carlos},
  journal   = {arXiv preprint arXiv:1604.06174},
  year      = {2016},
  url       = {https://arxiv.org/abs/1604.06174}
}

@misc{taori2023stanford,
  title={Stanford {Alpaca}: An Instruction-following {LLaMA} Model},
  author={Taori, Rohan and Gulrajani, Ishaan and Zhang, Tianyi and Dubois, Yann and Li, Xuechen and Guestrin, Carlos and Liang, Percy and Hashimoto, Tatsunori B},
  year={2023},
  url={https://github.com/tatsu-lab/stanford_alpaca}
}

@article{clark2018think,
  title={Think you have Solved Question Answering? {Try} {ARC}, the {AI2} Reasoning Challenge},
  author={Clark, Peter and Cowhey, Isaac and Etzioni, Oren and Khot, Tushar and Sabharwal, Ashish and Schoenick, Carissa and Tafjord, Oyvind},
  journal={arXiv preprint arXiv:1803.05457},
  year={2018}
}

@inproceedings{zellers2019hellaswag,
  title={{HellaSwag}: Can a Machine Really Finish Your Sentence?},
  author={Zellers, Rowan and Holtzman, Ari and Bisk, Yonatan and Farhadi, Ali and Choi, Yejin},
  booktitle={Proceedings of the 57th Annual Meeting of the Association for Computational Linguistics (ACL)},
  pages={4791--4800},
  year={2019}
}

@article{hendrycks2020measuring,
  title={Measuring massive multitask language understanding},
  author={Hendrycks, Dan and Burns, Collin and Basart, Steven and Zou, Andy and Mazeika, Mantas and Song, Dawn and Steinhardt, Jacob},
  journal={arXiv preprint arXiv:2009.03300},
  year={2020}
}

@article{gao2021framework,
  title={A framework for few-shot language model evaluation},
  author={Gao, Leo and Tow, Jonathan and Biderman, Stella and Black, Sid and DiPofi, Anthony and Foster, Charles and Golding, Laurence and Hsu, Jeffrey and McDonell, Kyle and Muennighoff, Niklas and others},
  journal={Zenodo},
  year={2021}
}

@inproceedings{lv2024lomo,
    title = "Full Parameter Fine-tuning for Large Language Models with Limited Resources",
    author = "Lv, Kai  and
      Yang, Yuqing  and
      Liu, Tengxiao  and
      Guo, Qipeng  and
      Qiu, Xipeng",
    editor = "Ku, Lun-Wei  and
      Martins, Andre  and
      Srikumar, Vivek",
    booktitle = "Proceedings of the 62nd Annual Meeting of the {A}ssociation for {C}omputational {L}inguistics (Volume 1: Long Papers)",
    month = aug,
    year = "2024",
    address = "Bangkok, Thailand",
    publisher = "Association for Computational Linguistics",
    url = "https://aclanthology.org/2024.acl-long.445/",
    doi = "10.18653/v1/2024.acl-long.445",
    pages = "8187--8198"
}

@inproceedings{paszke2019pytorch,
      title={{PyTorch}: An Imperative Style, High-Performance Deep Learning Library}, 
      author={Adam Paszke and Sam Gross and Francisco Massa and Adam Lerer and James Bradbury and Gregory Chanan and Trevor Killeen and Zeming Lin and Natalia Gimelshein and Luca Antiga and Alban Desmaison and Andreas Köpf and Edward Yang and Zach DeVito and Martin Raison and Alykhan Tejani and Sasank Chilamkurthy and Benoit Steiner and Lu Fang and Junjie Bai and Soumith Chintala},
      year={2019},
      booktitle={Advances in Neural Information Processing Systems (NeurIPS)},
      eprint={1912.01703},
      archivePrefix={arXiv},
      primaryClass={cs.LG},
      url={https://arxiv.org/abs/1912.01703}, 
}

@inproceedings{loshchilov2019adamw,
      title={Decoupled Weight Decay Regularization}, 
      author={Ilya Loshchilov and Frank Hutter},
      year={2019},
      booktitle={International Conference on Learning Representations (ICLR)},
      eprint={1711.05101},
      archivePrefix={arXiv},
      primaryClass={cs.LG},
      url={https://arxiv.org/abs/1711.05101}, 
}

@misc{granite2024,
  title={Granite 3.0 Language Models},
  url={https://github.com/ibm-granite/granite-3.0-language-models/},
  author={{IBM Granite Team}},
  month={October},
  year={2024}
}

@inproceedings{lv2024adalomo,
      title={{AdaLomo}: Low-memory Optimization with Adaptive Learning Rate}, 
      author={Kai Lv and Hang Yan and Qipeng Guo and Haijun Lv and Xipeng Qiu},
      year={2024},
      booktitle={Findings of the Association for Computational Linguistics: ACL 2024},
      pages={12486--12502},
      eprint={2310.10195},
      archivePrefix={arXiv},
      primaryClass={cs.LG},
      url={https://arxiv.org/abs/2310.10195}, 
}

@article{liu2025revffn,
      title={{RevFFN}: Memory-Efficient Full-Parameter Fine-Tuning of Mixture-of-Experts {LLMs} with Reversible Blocks}, 
      author={Ningyuan Liu and Jing Yang and Kaitong Cai and Keze Wang},
      year={2025},
      journal={arXiv preprint arXiv:2512.20920},
      eprint={2512.20920},
      archivePrefix={arXiv},
      primaryClass={cs.LG},
      url={https://arxiv.org/abs/2512.20920}, 
}

@inproceedings{dettmers2022optimizers,
      title={8-bit Optimizers via Block-wise Quantization}, 
      author={Tim Dettmers and Mike Lewis and Sam Shleifer and Luke Zettlemoyer},
      year={2022},
      booktitle={International Conference on Learning Representations (ICLR)},
      eprint={2110.02861},
      archivePrefix={arXiv},
      primaryClass={cs.LG},
      url={https://arxiv.org/abs/2110.02861}, 
}

@inproceedings{he2022fastermoe,
  title={{FasterMoE}: Modeling and Optimizing Training of Large-Scale Dynamic Pre-Trained Models},
  author={He, Jiaao and Zhai, Jidong and Antunes, Tiago and Wang, Haojie and Luo, Fuwen and Shi, Shangfeng and Li, Qin},
  booktitle={Proceedings of the 27th ACM SIGPLAN Symposium on Principles and Practice of Parallel Programming (PPoPP)},
  pages={120--134},
  year={2022}
}

@article{li2020pytorch,
  title={{PyTorch} Distributed: Experiences on Accelerating Data Parallel Training},
  author={Li, Shen and Zhao, Yanli and Varma, Rohan and Salpekar, Omkar and Noordhuis, Pieter and Li, Teng and Paszke, Adam and Smith, Jeff and Vaughan, Brian and Damania, Pritam and others},
  journal={arXiv preprint arXiv:2006.15704},
  year={2020}
}

@inproceedings{ren2021zero,
  title={{ZeRO-Offload}: Democratizing Billion-Scale Model Training},
  author={Ren, Jie and Rajbhandari, Samyam and Aminabadi, Reza Yazdani and Ruwase, Olatunji and Yang, Shuangyan and Zhang, Minjia and Li, Dong and He, Yuxiong},
  booktitle={2021 USENIX Annual Technical Conference (USENIX ATC 21)},
  pages={551--564},
  year={2021}
}

@article{yu2024collage,
  title={Collage: Light-Weight Low-Precision Strategy for {LLM} Training},
  author={Yu, Tao and Gupta, Gaurav and Gopalswamy, Karthick and Mamidala, Amith and Zhou, Hao and Huynh, Jeffrey and Park, Youngsuk and Diamant, Ron and Deoras, Anoop and Huan, Luke},
  journal={arXiv preprint arXiv:2405.03637},
  year={2024}
}

@article{hu2021lora,
  title={{LoRA}: Low-Rank Adaptation of Large Language Models},
  author={Hu, Edward J and Shen, Yelong and Wallis, Phillip and Allen-Zhu, Zeyuan and Li, Yuanzhi and Wang, Shean and Wang, Lu and Chen, Weizhu},
  journal={arXiv preprint arXiv:2106.09685},
  year={2021}
}

@inproceedings{dettmers2023qlora,
  title={{QLoRA}: Efficient Finetuning of Quantized {LLMs}},
  author={Dettmers, Tim and Pagnoni, Artidoro and Holtzman, Ari and Zettlemoyer, Luke},
  booktitle={Advances in Neural Information Processing Systems (NeurIPS)},
  pages={10088--10115},
  year={2023}
}

@inproceedings{wang2024let,
  title={Let the expert stick to his last: Expert-specialized fine-tuning for sparse architectural large language models},
  author={Wang, Zihan and Chen, Deli and Dai, Damai and Xu, Runxin and Li, Zhuoshu and Wu, Yu},
  booktitle={Proceedings of the 2024 Conference on Empirical Methods in Natural Language Processing},
  pages={784--801},
  year={2024}
}

@article{zhao2024galore,
  title={{GaLore}: Memory-Efficient {LLM} Training by Gradient Low-Rank Projection},
  author={Zhao, Jiawei and Zhang, Zhenyu and Chen, Beidi and Wang, Zhangyang and Anandkumar, Anima and Tian, Yuandong},
  journal={arXiv preprint arXiv:2403.03507},
  year={2024}
}

@inproceedings{luo2024badammemoryefficientparameter,
      title={{BAdam}: A Memory Efficient Full Parameter Optimization Method for Large Language Models}, 
      author={Qijun Luo and Hengxu Yu and Xiao Li},
      year={2024},
      booktitle={Advances in Neural Information Processing Systems (NeurIPS)},
      eprint={2404.02827},
      archivePrefix={arXiv},
      primaryClass={cs.LG},
      url={https://arxiv.org/abs/2404.02827}, 
}

@inproceedings{pascanu2013difficultytrainingrecurrentneural,
      title={On the difficulty of training Recurrent Neural Networks}, 
      author={Razvan Pascanu and Tomas Mikolov and Yoshua Bengio},
      year={2013},
      booktitle={Proceedings of the 30th International Conference on Machine Learning (ICML)},
      eprint={1211.5063},
      archivePrefix={arXiv},
      primaryClass={cs.LG},
      url={https://arxiv.org/abs/1211.5063}, 
}

@article{kukreja2026forge,
  title={{FORGE}: Fused On-Register Gradient Elimination for Memory-Efficient {LLM} Training},
  author={Kukreja, Dikshant and Prasad, Kritarth and Anand, Avinash and Wang, Zhengkui and Cambria, Erik and Liu, Timothy and Ng, Aik Beng and See, Simon and Chatterjee, Bapi},
  journal={arXiv preprint arXiv:2606.22932},
  year={2026}
}



\end{document}